\documentclass[sn-mathphys]{sn-jnl}

\usepackage{indentfirst} 
\usepackage{array}
\usepackage{booktabs}
\usepackage{caption}  
\usepackage{graphicx}

\jyear{2022}%

\theoremstyle{thmstyleone}%
\theoremstyle{thmstyletwo}%

\theoremstyle{thmstylethree}%

\begin{document}

\title[Article Title]{CoT-MISR:Marrying Convolution and Transformer for Multi-Image Super-Resolution}


\author{\fnm{Mingming} \sur{Xiu}}

\author{\fnm{Yang} \sur{Nie}}

\author{\fnm{Qing} \sur{Song}}

\author{\fnm{Chun} \sur{Liu}}






\abstract{\par As a method of image restoration, image super-resolution has been extensively studied at first. How to transform a low-resolution image to restore its high-resolution image information is a problem that researchers have been exploring. In the early physical transformation methods, the high-resolution pictures generated by these methods always have a serious problem of missing information, and the edges and details can not be well recovered. With the development of hardware technology and mathematics, people begin to use in-depth learning methods for image super-resolution tasks, from direct in-depth learning models, residual channel attention networks, bi-directional suppression networks, to tr networks with transformer network modules, which have gradually achieved good results. In the research of multi-graph super-resolution, thanks to the establishment of multi-graph super-resolution dataset, we have experienced the evolution from convolution model to transformer model, and the quality of super-resolution has been continuously improved. However, we find that neither pure convolution nor pure tr network can make good use of low-resolution image information. Based on this, we propose a new end-to-end CoT-MISR network. CoT-MISR network makes up for local and global information by using the advantages of convolution and tr. The validation of dataset under equal parameters shows that our CoT-MISR network has reached the optimal score index.}




\maketitle

\section{Introduction}\label{sec1}

Image supersection is a technique to restore image details in image processing. The early image supersection methods are mainly based on physical interpolation \cite{1} methods and image reconstruction \cite{2,3,4} methods. These methods are only explicit expansion for the restoration of image details, and can not restore texture details well. After that, researchers also explored many methods in the field of image super segmentation. Although the performance of super segmentation has been improved, it still fails to achieve satisfactory results.

The development of deep learning has shifted researchers' attention to the direction of neural networks. The emergence of single map super-resolution neural networks \cite{5} has led to new exploration in the field of image superdivision. Residual attention mechanism \cite{4}, two-way regression inhibition \cite{6,7,8,9,10}, deep convolution \cite{7}, generation confrontation \cite{11,12,13,14} and other networks have been put forward successively. These new ideas have reference significance in the field of image superdivision. Later, transformer \cite{15,16} \cite{39,40,41} was introduced to supplement the global semantic information of the image, but the role of convolutional network in fusing local information was also ignored to some extent.

The progress in the field of single image super segmentation has become the basis of image super segmentation technology in the field of remote sensing. In the field of remote sensing, it is mainly to restore the high-definition image of a certain region on the basis of multiple images. The main technical difficulty is how to align and fuse multiple images to restore high-definition remote sensing images of the region due to different acquisition times, locations and weather. The proposal of multigraph super-resolution convolution network shows the advantages of deep learning in the field of remote sensing multigraph super-resolution. The MISR convolution neural network \cite{17,18,19,20,21} and the migrated transformer \cite{16} have improved their performance in multigraph super-resolution. However, in order to improve the quality of super-resolution, the number of network parameters increases. The convolutional network focuses on local information, while the transformer focuses on global information. None of the above multiimage super-resolution networks can make good use of local information and global information of low resolution images at the same time.

We use the data provided by PROBA-V \cite{22} for training and propose an optimal end-to-end multigraph super-resolution network. Our main innovations are as follows:

1) A new multi image super-resolution network block structure is proposed to improve the use efficiency of low resolution image information and improve the full fusion of local and global information of low resolution image;

2) The new network structure reduces the number of parameters by changing the convolution usage structure, and at the same time, the details feature extraction is more delicate, achieving the optimal result in the test set divided by PROBA-V;

3) Compared with the current optimal convolution structure and transformer structure, this network has significant advantages in the field of multigraph superresolution, and it is also useful for feature extraction in other fields.

\section{Related Work}\label{sec2}
\subsection{Single Image Super-Resolution}\label{subsec2-1}

The super-resolution task was initially applied to the super-resolution of single image. The most important part of convolutional neural network is the feature extraction module. The proposed end-to-end network mainly uses the network composed of convolutional blocks to map and learn features, which is inefficient in feature extraction. The proposed deep convolution network \cite{7} enhances the ability of information extraction, but the increase of parameter quantity also leads to the enhancement of noise influence, and the improvement of super resolution effect is limited. The residual channel attention network module \cite{4} is proposed. By using the residual and channel attention mechanism, the fusion of front and back feature information and the extraction of key feature information are considered, and the super-resolution effect is improved. The proposed dual stream suppression network \cite{10} reduces the effect of super-resolution one to many mapping, but also reduces the quality of super-resolution to a certain extent. The proposed generation countermeasure network \cite{13,14} is close to the essence of super division. It generates more abundant pixel information based on the existing pixel information, but there is still a shortage of feature generation for the super resolution of natural images.

\subsection{Multi-image Super-resolution}\label{subsec2-2}

At present, most multigraph super-resolution methods based on depth learning use the encoder decoder structure to complete the encoding, fusion and decoding of feature information. Multi image super-resolution will obtain a number of low resolution images after encoding and input them into the feature extraction network module, and then obtain high resolution images after feature decoding.

Tsai \cite{23} was the first to conduct research in the field of multi image superresolution, and used frequency domain technology to improve the spatial resolution of images by combining multiple images with sub-pixel displacement. Since the frequency shift method has some defects in merging the prior information of HR images, several spatial domain super division techniques \cite{24} have emerged, including convex set projection (POCS) \cite{25}, non-uniform interpolation \cite{26}, regularization methods \cite{27,28} and sparse coding \cite{29}.

In recent years, deep learning-based methods have been used to solve the hyperresolution task \cite{30,31,32} of video image enhancement. Kawulok \cite{33} and others used the learning-based SISR method using EvoNet framework \cite{34} based on several deep CNNs to utilize SISR in the pre-processing phase of MISR input data.

Molini\cite{35} et al. proposed a new architecture based on CNN, DeepSUM, using the remote sensing dataset PROBA-V as the benchmark. By utilizing spatial and temporal correlation, an end-to-end learning method was established. Deudon \cite{18} et al. proposed HighRes network based on deep learning. Inspired by video superresolution 3DSRNet \cite{36}, Francisco Dorr \cite{37} et al.  proposed a 3D WSRNet that uses WDSR blocks to obtain temporal correlation between frames. WDSR-MFSR \cite{38} uses multiple WDSR residual blocks to further enhance feature extraction. Francesco Salvetti \cite{20} et al. proposed RAMS. RAMS avoids the effect of time series on feature fusion. Then, An \cite{16} et al. introduced the transformer module into TR-MISR, and used transformer to extract global information and self-attention mechanism to get good overscores.

At present, the existing multi-graph super-resolution network puts forward a new idea of super-resolution after summarizing the shortcomings of the network before and referring to the advantages of the mainstream task network module. However, there are still some deficiencies in the utilization of local and global information in low-resolution images, and detailed texture generation can not achieve good results.





\section{Methodology}\label{sec3}

\begin{figure}[ht]
	\begin{center}
		\includegraphics[width=1\linewidth]{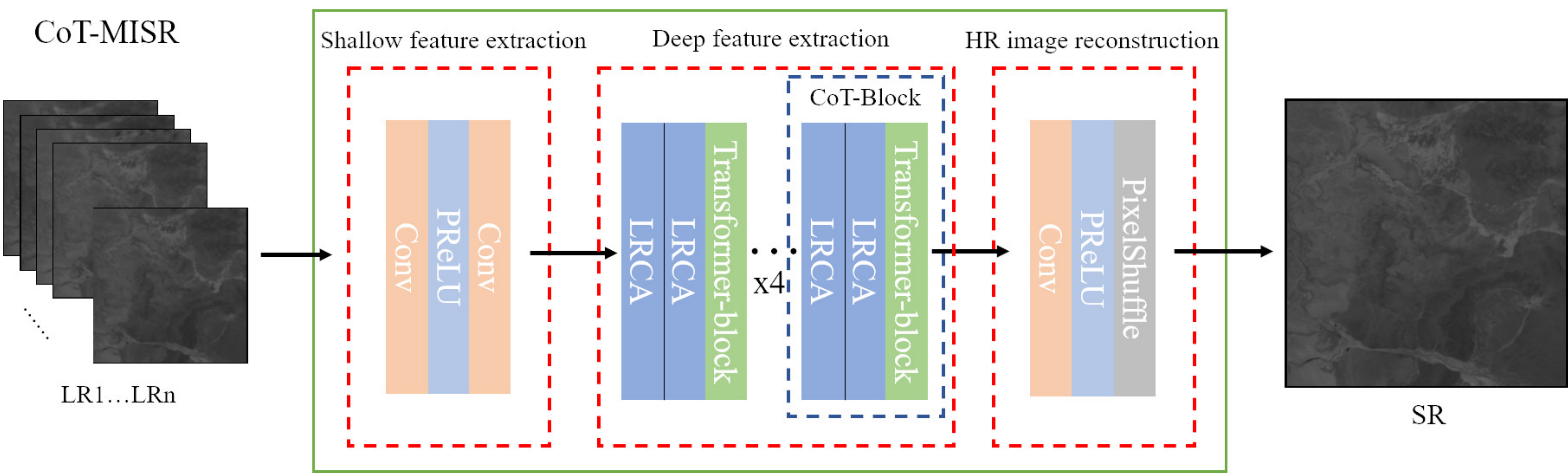}
	\end{center}
	\caption{Overview of CoT-MISR:Based on the characteristics of MISR, we preprocess the input images of the same area, and uniformly adjust the channels of the input low-resolution images to a single value. The low-resolution images are recorded as LR$_{1}-LR_{k}$ and the k value can be set manually. The images that needs to be learned is obtained from the shallow feature extraction module. We send the feature maps to the deep feature extraction module for feature fusion and transfer the learned features to the image reconstruction network to generate a high-resolution image SR for the same area.}
	\label{cot}
\end{figure}

In this chapter, we will introduce the main network structure of the proposed CoT-MISR method, as shown in figure \ref{cot} .It also introduces the light residual channel attention module (LRCA) and transformer module (T-Block) which compose the CoT-MISR network.

\subsection{Shallow feature extraction}\label{subsec3-1}

We use size B×H×W×K×C$_{i}$ tensor to represent the size of low resolution images $\lbrace LR \rbrace _{i=1} ^{k}$. Where B, H, W, K and C$_{i}$ represent the input batch size, image height, image width, and image number and the number of image channels (C$_{i}$=1). $\lbrace LR \rbrace _{i=1} ^{k}$, where k can only be set manually. If the number of input images is n and n $<$ k, (k $-$ n) will be generated to complement the image alignment k. Because the time environment of multiple input images is different, there is information imbalance of different degrees. We use the Median function to calculate the reference image LRLR$_{ref}$ of an image sequence to solve the problem that the information of multiple images in the MISR task is greater than that of any image, as shown in EQ \ref{eq1}.

\begin{equation}
LR_{ref} = Median(LR _{1}, . . . , LR _{k})
\label{eq1}
\end{equation}

After Median operation reference images $\lbrace LRref \rbrace $ and $\lbrace LR \rbrace _{i=1} ^{k}$ are combined as G$_{i}$.

\begin{equation}
G_{i} = [LR_{ref}, LR_{i}]\mid_{i=1} ^{k} 
\label{eq2}
\end{equation}

In the shallow feature extraction we will perform the reshape operation on the input, followed by two layers of convolution, and obtain the learning characteristic graph Br $\in\mathbb{R}^{H \times W \times Ce}$ with the number of channels Ce through the convolution operation.Among them, C$_{e}$ $>$ C$_{i}$ can better extract features in the feature fusion process by expanding the number of channels. The operation can be represented by Eq 3, and the shallow feature extraction structure is shown in Fig \ref{cot}.

\begin{equation}
br=shallow(G_{i}\mid_{i=1}^{k})
\label{eq3}
\end{equation}

The whole network can be expressed as:

\begin{equation}
sr = Reconstruction(CoT(Conv(Median(LR_{i=1} ^{k})))
\label{eq4}
\end{equation}

Where CoT and Reconstruction represent CoT-Block blocks and the HR image reconstruction blocks.Conv represents convolution operation and Reconstruction represents HR image reconstruction operation.

\subsection{Deep feature extraction}\label{subsec3-2}
\subsubsection{Light residual channel attention module(LRCA)}\label{subsub3-2-1}

The residual structure can enhance the stability of network data transmission and improve the learning performance of the network. The channel attention structure can enhance the fitting of channel data with obvious features. On the basis of the residual channel attention structure, we have changed the structure of feature preprocessing convolution. The introduction of depth separable convolution reduces the number of parameters, while enhancing the spatial attention of low resolution images. Eq \ref{eq5} represents the formula of the LRCA module.

\begin{equation}
lrca=CA(SA(br)) + br
\label{eq5}
\end{equation}

\begin{figure}[t]
	\begin{center}
		\includegraphics[width=0.7\linewidth]{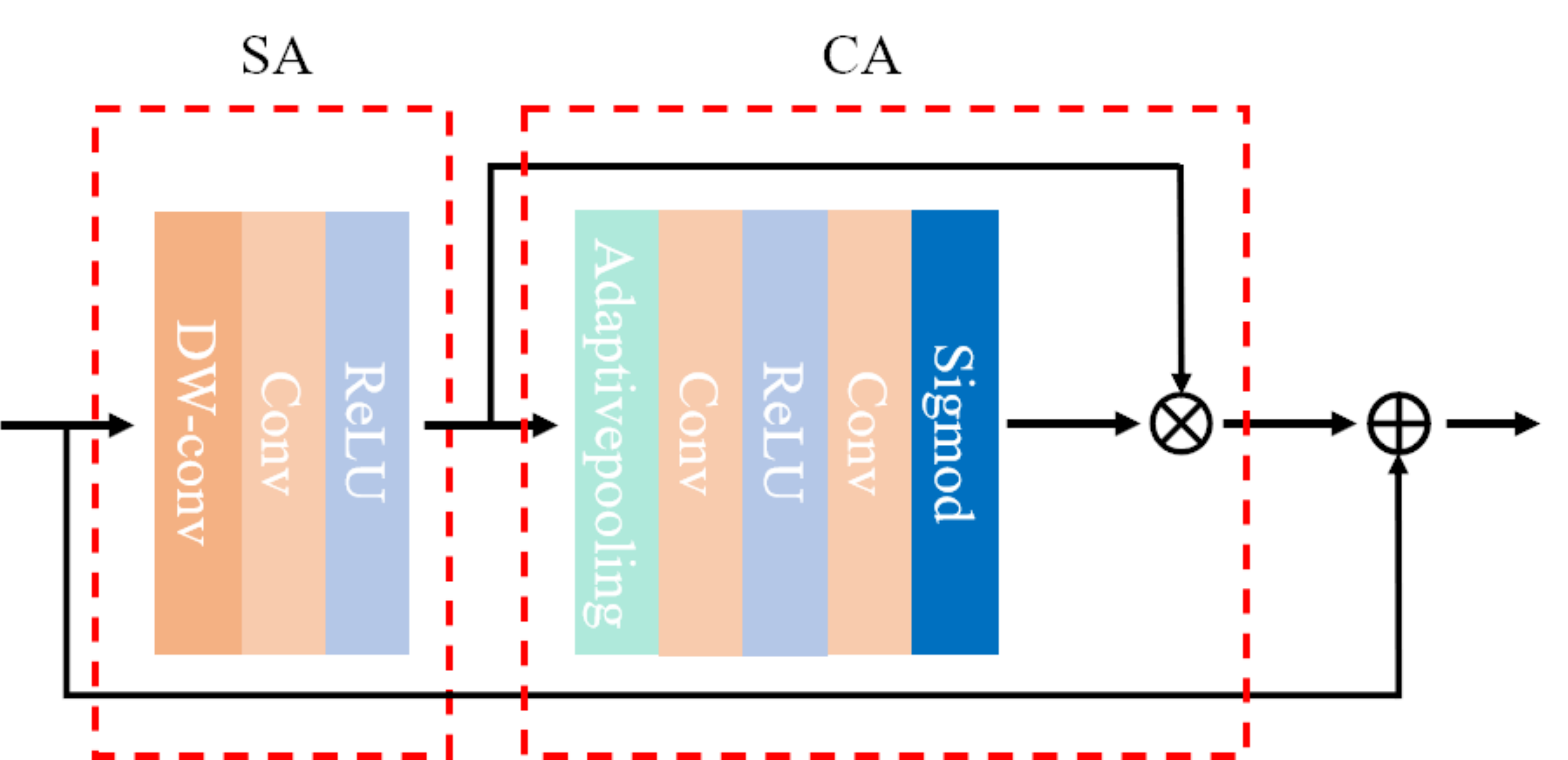}
	\end{center}
	\caption{Overall block diagram of light residual channel attention (LRCA) module. We divide it into convolutional spatial attention part (SA) and channel attention part (CA).}
	\label{lrca}
\end{figure}

Where SA represents the spatial attention part of light volume convolution, and CA represents the channel attention part based on convolution. The specific structure of LRCA module is shown in Fig \ref{lrca}.

\subsubsection{T-Block}\label{subsubsec3-2-2}

\begin{figure}[ht]
	\begin{center}
		\includegraphics[width=0.7\linewidth]{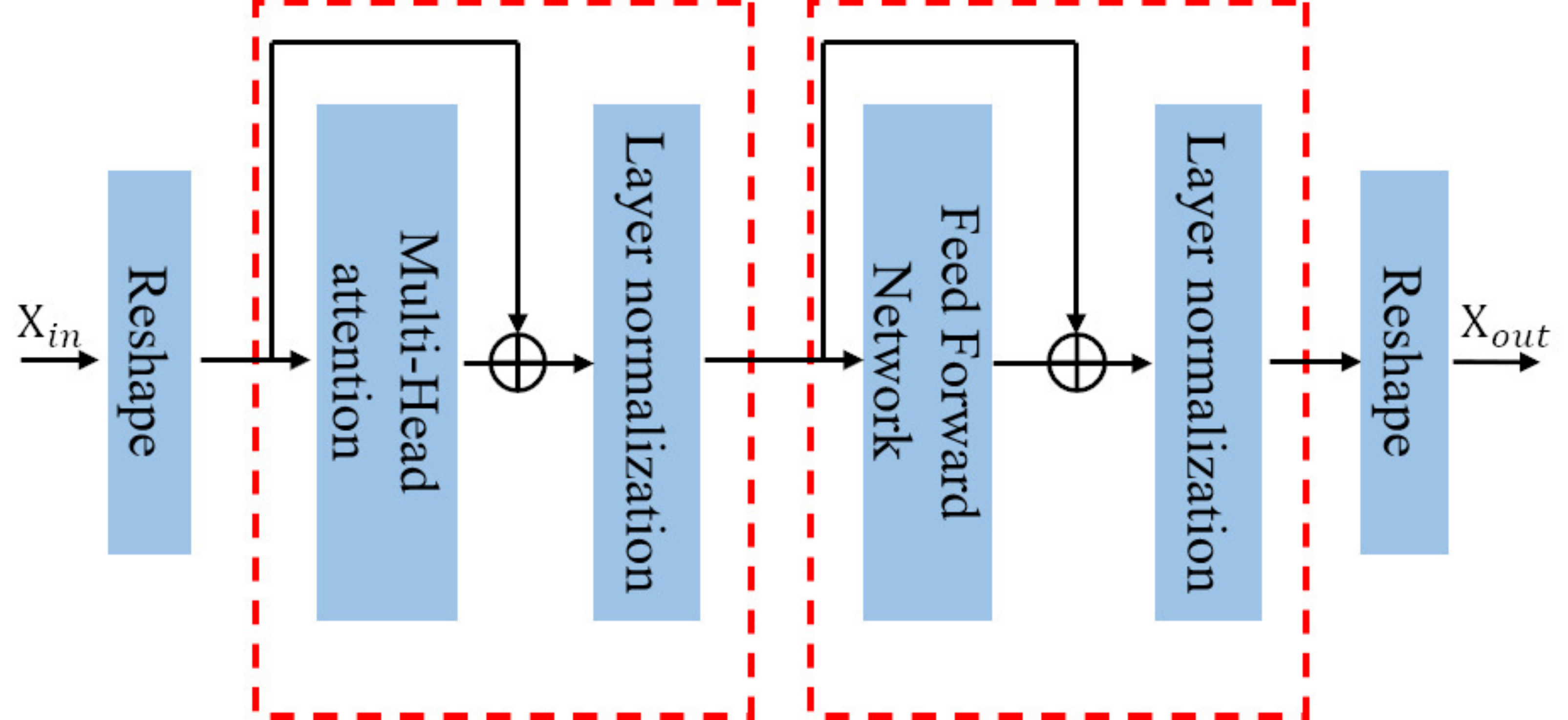}
	\end{center}
	\caption{Set the input tensor X$_{in}$ size to B × C × H × W. After reshape (H × W) × B × C. The output tensor X$_{out}$ is restored to B × C × H × W after reshape operation.}
	\label{t-block}
\end{figure}

In the new network, we deconstruct the standard six layer transformer structure, and only take four layers as the global learning network module. While reducing the number of parameters, we make full use of the local information transmitted by the lightweight residual attention module to learn the global information, which enhances the ability to learn network details. Fig \ref{t-block} shows the structure diagram of the transformer module we changed. We perform the reshape operation at the beginning and end of the transformer destruction layer to achieve the goal of fusion training with convolution.

The main calculation process of T-Block module is as follows:

\begin{equation}
	tr=Tr\lbrace reshape1(lrca)\rbrace 
	\label{eq6}
\end{equation}

\begin{equation}
bi = reshape2(tr)
\label{eq7}
\end{equation}

Where B$_{i}\in\mathbb{R}^{H \times W \times Ce}$ is the intermediate tensor of convolution and transformer operations in the CoT-MISR module.

\subsubsection{CoT-Block}\label{subsubsec3-2-3}
CoT-Block module is composed of lightweight residual convolution and transformer layered module after deconstruction. The structure of CoT-Block module achieves the optimal result in multi graph super division. The feature fusion process of CoT-Block module is shown in EQ \ref{eq8}.

\begin{equation}
	CoT(i)= Tr\lbrace Lrca(bi))\rbrace
	\label{eq8}
\end{equation}

\subsection{HR image reconstruction}\label{subsec3-3}

In the deconvolution used in general super-resolution tasks, there will be a large number of regions for filling zeros, which may affect the final super division effect in the convolution calculation results. The single pixel on the feature can be combined into a unit on the feature by sub-pixel convolution. The pixel on each feature is equivalent to the sub-pixel on the new feature.
In order to achieve the size after superresolution, we use a layer of convolution and sub-pixel convolution operations to reach the HR scale of the original image. The convolution operation converts the characteristic graph CoT(i) into the number of channels of HR image $\in\mathbb{R}^{H \times W \times C}$. The PixelShuffle operation obtains SR images with the same tensor size as HR.

\section{Experiment}\label{sec4}
In this part, we mainly introduce relevant experiments, including experiment configuration, data set introduction, data changes in each layer of the training network process, experimental results analysis, etc.

\subsection{Proba-V Dataset}\label{subsec4-1}
The remote sensing satellite image dataset PROBA-V provided by the Advanced Concept Group of the European Space Agency (ESA) is the main multi map hyperspectral dataset in the field of remote sensing. PROBA-V data set is specially used for super-resolution research of remote sensing images. The data set is taken by proba-v satellite and is divided into two kinds of spectral data RED and NIR. The satellite takes different resolutions to photograph the terrain at the same position, location and height in different time periods. The dataset consists of 128 $\times$ 128 low resolution images and 384 $\times$ 384 high resolution images. We use TR-MISR's preprocessing method for PROBA-V dataset to optimize the data structure, mask the low resolution images of each part according to the original data information, and improve the image information quality of training and testing.
We align the input data and extract the information through preprocessing. We divide the data set into training set and verification set according to the ratio of 9:1 for experiment.

\subsection{Experimental parameters}\label{subsec4-2}

The experiment was carried out in an Ubuntu 18.04 server operating system based on 64 bit Linux. The server is configured with 4 Nvidia 1080ti GPU (12 GB memory) and Intel (R) Core (TM) i7-6700K CPU @ 4.00GHz. We use the cPSNR and cSSIM proposed by ESA for PROBA-V dataset as evaluation indicators, which can well show the closeness between SR pixel values and HR actual pixel values.

In the training network, we set two learning rates, the initial coding network learning rate is 0.002, and the initial feature fusion cot module learning rate is 0.001. Enter the batch size as 8. The training period of the part of the model containing transformer is long. We set the training period as 100 epochs in the ablation phase and 400 epochs in the indicator verification phase. In the model ablation experiment, we used NIR type data in the dataset and verified the results. In the indicator verification stage, the verification is conducted on NIR, RED and NIR\&RED respectively.

\begin{figure}[ht]
	\begin{center}
		\includegraphics[width=0.7\linewidth]{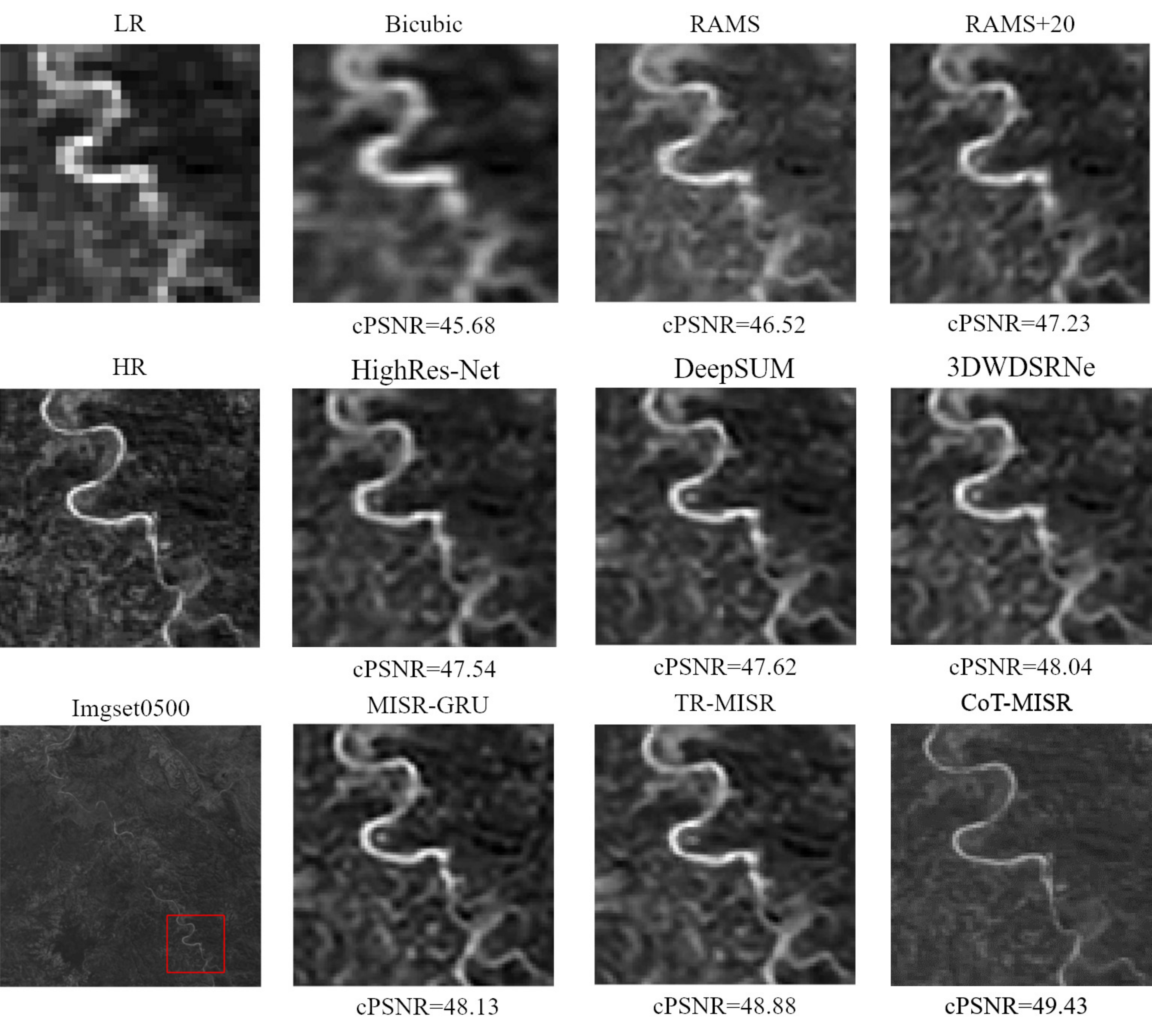}
	\end{center}
	\caption{This figure represents a comparison of local performance between different MISR methods on the imgset0500 scene of the RED band, based on the cPSNR evaluation metric.}
	\label{cpsnr}
\end{figure}

\subsection{Comparison}\label{subsec4-3}
We take PROBA-V as the data set and divide the training set with a ratio of 9:1. Given the same training/verification set, we choose representative methods in the super-resolution field and compare them in single band data NIR/RED and full band data ALL. The evaluation indicators include cPSNR and cSSIM.

This paper briefly introduces various methods and their experimental settings in the super-resolution field compared with CoT-MISR.

\par1) Bicubic: This is the baseline method, selecting the clearest image in each scene and performing bicubic interpolation. 
\par2) RCAN: It proposes channel focus (CA) to deal with different channels. The experiment sets the residual group and residual channel to 5.
\par3) VSR-DUF: It uses dynamic upsampling filters to generate corresponding filters for different inputs. The experiment sets the number of input video frames to 9 and selects a 16-layer framework.
\par4) IBP: It is one of the most classical algorithms for image super-resolution, which improves the resolution of an image through iterations. In the experiment, bicubic interpolation was used to obtain the initial solution, and phase correlation algorithm was used for registration.
\par5) BTV: It is an image enhancement method that focuses on restoring image edges and removing noise in super-resolution tasks. It minimizes the l1 norm as the loss in each iteration, while using bilateral regularization.
\par6) HighRes-Net: This method, which won the runner-up in the PROBA-V Challenge, is an end-to-end framework that combines an encoder-decoder network and a registration network for joint training. The experiment used the default framework and set the number of input images to 16.
\par7) MISR-GRU: It uses convGRU to fuse different features and obtain the fused feature by processing the hidden states. The number of input images in the experiment is set to 24.
\par8) DeepSUM: This network, which won the PROBA-V challenge, is a deep framework that focuses on exploring the spatiotemporal correlations between images. The experiment used 9 images. The authors also released DeepSUM++ which improved the model by introducing graph convolution in the encoder.
\par9) 3DWDSRNet: It emphasizes the acquisition of temporal variations between frames, and the experiment sets the number of input images to 7.
\par10) RAMS: It is currently the state-of-the-art MISR method on the PROBA-V Kelvin dataset, based on a large number of manually designed attention blocks. The experiment sets the number of input images to 9. RAMS+20 adopts temporal self-ensemble, which randomly shuffles the input image sequence 20 times to obtain the average image at the expense of reduced inference speed.
\par11)TR-MISR: It is a network that first introduced the transformer structure in multi-image super-resolution and addressed the fusion problem in multi-image super-resolution in the super-resolution structure. It made major representations in tensor transformation and training methods and achieved the best results in multi-image super-resolution. However, the corresponding parameter and computational complexity are quite large.

Table \ref{tab1} demonstrates that our proposed CoT-MISR network structure achieves the best performance in terms of cPSNR and cSSIM evaluation metrics compared to the previous convolutional and transformer models on the NIR, RED, and ALL validation sets. 

\begin{table}[h]
\centering
\caption{\centering{Result comparison of MISR models}}\label{tab1}			
\begin{tabular}{lcccccc}
\toprule
\multirow{2}*{Method} & \multicolumn{2}{c}{NIR} & \multicolumn{2}{c}{RED} & \multicolumn{2}{c}{ALL} \\
\cmidrule(lr){2-3} \cmidrule(lr){4-5} \cmidrule(lr){6-7}
& cPSNR & cSSIM & cPSNR & cSSIM & cPSNR & cSSIM \\
\midrule
Bicubic    	& 45.44   & 0.9770  & 47.33  & 0.9840  & 46.40  & 0.9806 \\
BTV    		& 45.93   & 0.9794  & 48.12  & 0.9861  & 47.04  & 0.9828 \\
IBP    		& 45.96   & 0.9796  & 48.21  & 0.9865  & 47.10  & 0.9831 \\
RCAN    	& 45.66   & 0.9798  & 48.22  & 0.9870  & 46.96  & 0.9835 \\
VSR-DUF    	& 47.20   & 0.9850  & 49.59  & 0.9902  & 48.42  & 0.9876 \\
HighRes-Net & 47.55   & 0.9855  & 49.75  & 0.9904  & 48.67 	& 0.9880 \\
3DWDSRNet   & 47.58   & 0.9856  & 49.90  & 0.9908  & 48.76  & 0.9882 \\
DeepSUM++   & 47.84   & 0.9858  & 50.00  & 0.9908  & 48.94  & 0.9883 \\
MISR-GRU    & 47.88   & 0.9861  & 50.11  & 0.9910  & 49.01  & 0.9886 \\
DeepSUM++   & 47.93   & 0.9862  & 50.08  & 0.9912  & 49.02  & 0.9887 \\
RAMS    	& 47.17   & 0.9869  & 50.13  & 0.9910  & 49.17  & 0.9890 \\
RAMS$_{+20}$ & 48.27  & 0.9870  & 50.27  & 0.9912  & 49.29  & 0.9891 \\
TR-MISR    	& 48.54   & 0.9882  & 50.67  & 0.9921  & 49.62  & 0.9902 \\
CoT-MISR$_{ours}$ & \textbf{51.41} & \textbf{0.9914} & \textbf{52.86} & \textbf{0.9965} & \textbf{52.03} & \textbf{0.9941} \\
\botrule
\end{tabular}			
\footnotetext{Performance of different methods on the validation set, where the bold entities represent the best results for different evaluation metrics within the same dataset.}
\end{table}
 
\subsection{Ablation experiment}\label{subsec4-4}

\subsubsection{CoT-MISR's structure}\label{subsubsec4-4-1}

\par We have proposed three structures, namely, 8-layer LRCA\&4-layer T-Block structure, 2-layer LRCA\&1-layer T-Block structure, 4 groups, and 4-layer LRCA\&4-layer T-Block\&4-layer LRCA structure. Table \ref{tab2} records the experimental results of three experiments. We compare them on the NIR category test set. The number of training rounds is 100 epochs. The results show that the CoT-MISR structure achieves the best result in the combination of 2-layer of LRCA\&1-layer of T-Block.

\begin{table}[h]
	\centering
	\caption{\centering{CoT-MISR structure comparison results}}\label{tab2}			
	\begin{tabular}{cccccc}
	\toprule
	Architecture & Parameter & Dataset & Epochs & cPSNR & cSSIM \\
	\midrule
	8c4t 		& 235k & NIR & 100 & 50.14  & 0.9911 \\
	4c4t4c 		& 235k & NIR & 100 & 50.50  & 0.9911 \\
	$2c1t\times4$	& 235k & NIR & 100 & 50.59  & 0.9913 \\
	\botrule
	\end{tabular}			
	\footnotetext{In the table c stands for LRCA, t for T-Block, and numbers for number of modules.}
\end{table}

\subsubsection{Attention Ablation in Lightweight Channel Residual}\label{subsubsec4-4-2}

\par In order to verify the role of the light residual channel attention module in feature extraction, we carried out disassembly and ablation experiments on the light residual attention module, and successively compared the main performance results of the spatial channel attention (SCA), channel attention (CA), and spatial attention (SA) modules in the feature fusion process and after the combination of the transformer structure. Table 3 shows that the lightweight residual channel spatial attention module has significant advantages in obtaining spatial features and feature extraction. In the process of combining with the transformer structure, the advantages of convolution and transformer in structure complementarity are verified, and the optimal result of multi-image superresolution is achieved.

\begin{table}[h]
	\centering
	\caption{\centering{Ablation results of light residual channel attention module}}\label{tab3}			
	\begin{tabular}{ccccccc}
	\toprule
	CA & SA & Parameter & Dataset & Epochs & cPSNR & cSSIM \\
	\midrule
	\checkmark & \checkmark & 236k & NIR & 100 & 50.59 & 0.9913 \\
	\checkmark & - & 231k & NIR & 100 & 49.69  & 0.9901 \\
	- & \checkmark	& 210k & NIR & 100 & 49.81  & 0.9906 \\
	\botrule
	\end{tabular}			
	\footnotetext{The CA in the table represents the Channel Attention Module, the SA represents the Spatial Attention Module, and we process it lightly in the SA Module, \checkmark indicates that it contains the module.}
\end{table}

\section{Conclusion}\label{sce5}

This paper presents a new end-to-end framework for solving MISR tasks, cot. Inspired by residual channel attention, we combine the advantages of spatial attention in convoluted networks into a lightweight residual channel attention module. Use transformer's self-attention mechanism to compensate for convolution's drawbacks in global attention. The cot model combined with convolution and transformer further solves the problem of poor model adaptability and low data utilization in MISR tasks. The framework we propose mainly consists of three parts: input data precoding layer, feature fusion layer and decoding layer. Our experiment discussed the respective roles of spatial and channel attention in MISR tasks and structurally explored the optimal architecture of the MISR model. We compare cot with other existing MISR methods and show the advantages of our proposed cot module, which achieves the best results of current MISR methods.
\par Our proposed cot has reached the most advanced level on the PROBA-V dataset. The model combines the advantages of convolution and transformer, and improves performance while reducing the model parameters. The idea of convolution and transformer in low-level computer vision tasks has been broadened. The ablation experiments of cot fusion module show the importance of using Lr image pixel information effectively in the super-domain. The combination of space and global attention is a noteworthy research direction in super-resolution, and it is very helpful to improve image fusion in MISR.
\par Next, we will make further thinking from the following aspects: 1) Continue to study the lightweight residual channel attention module and make some attempts in spatial attention; 2) Continue to explore the combination and use of convolution and transformer; 3) Improve network input and validate it in other open datasets.







\bibliography{sn-bibliography}


\end{document}